\documentclass[10pt,twocolumn,letterpaper]{article}
\usepackage{iccv}
\usepackage{times}
\usepackage{epsfig}
\usepackage{graphicx}
\usepackage{amsmath}
\usepackage{amssymb}

\usepackage{caption}
\usepackage{csquotes}
\usepackage{multirow}
\usepackage{verbatim}
\usepackage{xcolor}
\usepackage{wrapfig}

\usepackage[pagebackref=true,breaklinks=true,letterpaper=true,colorlinks,bookmarks=false]{hyperref}

\iccvfinalcopy % *** Uncomment this line for the final submission

\def\iccvPaperID{1720} % *** Enter the ICCV Paper ID here

\ificcvfinal\fi

\DeclareMathOperator*{\argmax}{arg\,max}
\newcommand{\refsec}[1]{Sect.~\ref{#1}}
\newcommand{\reffig}[1]{Fig.~\ref{#1}}
\newcommand{\reftab}[1]{Tab.~\ref{#1}}
\newcommand{\refequ}[1]{$\left(\ref{#1}\right)$}

\usepackage{color}

\begin{document}

%%%%%%%%% TITLE
\title{Rotational Subgroup Voting and Pose Clustering for\\Robust 3D Object Recognition}

\author{Anders Glent Buch \quad Lilita Kiforenko \quad Dirk Kraft\\
University of Southern Denmark\\
{\tt\small \{anbu,lilita,kraft\}@mmmi.sdu.dk}
}

\maketitle
\thispagestyle{empty}

%%%%%%%%% ABSTRACT
\begin{abstract}
It is possible to associate a highly constrained subset of relative 6\,DoF poses between two 3D shapes, as long as the local surface orientation, the normal vector, is available at every surface point. Local shape features can be used to find putative point correspondences between the models due to their ability to handle noisy and incomplete data. However, this correspondence set is usually contaminated by outliers in practical scenarios, which has led to many past contributions based on robust detectors such as the Hough transform or RANSAC. The key insight of our work is that a single correspondence between oriented points on the two models is constrained to cast votes in a 1\,DoF rotational subgroup of the full group of poses, SE(3). Kernel density estimation allows combining the set of votes efficiently to determine a full 6\,DoF candidate pose between the models. This modal pose with the highest density is stable under challenging conditions, such as noise, clutter, and occlusions, and provides the output estimate of our method.

We first analyze the robustness of our method in relation to noise and show that it handles high outlier rates much better than RANSAC for the task of 6\,DoF pose estimation. We then apply our method to four state of the art data sets for 3D object recognition that contain occluded and cluttered scenes. Our method achieves perfect recall on two LIDAR data sets and outperforms competing methods on two RGB-D data sets, thus setting a new standard for general 3D object recognition using point cloud data.
\end{abstract}

%%%%%%%%% INTRO
\section{Introduction}
There is an ever-increasing need for robust perception systems and automated solutions in industry, service robotics and other applications. One of the great challenges is for an autonomous system to navigate in unstructured environments, which for manipulation tasks crucially relies on the ability to recognize and localize the parts or objects of interest. Although some recognition tasks naturally lend themselves to image-based techniques---some examples are pedestrian detection, traffic sign recognition and gesture recognition---it is vital for an autonomous agent to acquire the pose of objects in the ambient space to be able to perform any real manipulation tasks. This involves determining the full 3 DoF position and 3 DoF rotation of an object, which can become a computationally expensive operation in cluttered scenes.

Many contributions have been made on this matter \cite{guo20143d}, in recent years heavily based on range or 3D data in the form of either RGB-D images, point clouds or meshes. These data are acquired from Kinect sensors, industrial grade laser scanners or the likes. Steady improvements have been achieved, when considering overall object recognition performances in 3D data sets, where the aim usually is to find the full 6 DoF pose of multiple objects in unstructured or semi-structured scenes. In this very general free-form recognition and localization scenario, one needs to deal with some nuisances, including noise in the acquired sensor data and partial occlusions of the objects due to obscuring elements. In this work, we focus on the underlying problem of recovering the 6 DoF pose of an object under these conditions and adopt a feature-based approach, where multiple local shape features are used to describe a full 3D object model. These local features can be matched with a scene, providing a set of point correspondences between the object and the scene. In real applications this set of correspondences is heavily contaminated by outliers, making the search for the pose which brings the object into correct alignment with the scene a very challenging problem.

This paper describes a method that can be used to localize 3D objects within a scene acquired from a depth or 3D sensor. The method is shown to be particularly robust towards high fractions of outliers, which results in very competitive recognition rates for a number of applications. To achieve this, our method uses geometric constraints to cast full 6 DoF votes for the pose using individual correspondences. These votes are shown to lie on a small 1 DoF manifold, which allows for a tractable inference step based on kernel density estimation. Our method differs from previous methods since these usually require two \cite{drost2010model}, three \cite{aldoma2016global} or more correspondences to compute pose candidates, making the sampling process much more expensive. Our method uses single correpondences to vote for a set of candidate poses and delays the determination of the correct pose to a subsequent clustering process. We tested our method on two well-known free-form object recognition data sets and two recent RGB-D data set. In all cases, our method outperforms competing methods.

This paper is structured as follows. In \refsec{related} we provide an overview of related work within the field of object recognition and pose estimation in 3D data. \refsec{method} gives the details of our algorithm and in \refsec{recognition} we explain how our algorithm is used in a 3D object recognition and pose estimation pipeline. In \refsec{results} we present experimental results and in \refsec{conclusions} we conclude on our findings.

%%%%%%%%% RELATED
\section{Related work}\label{related}
Object recognition and pose estimation in 3D data has been an active research area for more than two decades. Early works include \cite{chua1997point,johnson1999using}, from which the well-known Spin Images used local shape descriptors and correspondence grouping for recognizing objects in range images. In the years that followed, several variations of local shape descriptors appeared. In \cite{frome2004recognizing} a local 3D shape context descriptor was used for matching segmented point cloud models. Similarly, in \cite{chen20073d} a set of local descriptors were used to recognize shapes in scenes with no or limited amounts of clutter. A more elaborate system for both 3D object modeling and recognition in cluttered scenes was presented in \cite{mian2006three}. Recognition was performed using randomly sampled point pairs for which an area-based descriptor was computed. The use of point pairs eased the process of computing a relative pose between the object and scene models but came at an increased computational cost. Progress in these cluttered scenarios continued, with other methods, \eg \cite{taati2011local}, using RANSAC \cite{fischler1981random} for robust pose estimation, and \cite{bariya20123d} using a tree search through the set of possible correspondences. In an influential work \cite{drost2010model}, the use of point pairs was revisited, but now with a simpler and computationally cheaper feature and a fast pose estimation algorithm using a variant of geometric hashing. The method was further developed in several later works, including \cite{birdal2015point,drost20123d,hinterstoisser2016going}. The development of local shape features continued for a considerable variation of applications such as mesh based keypoint detection and description \cite{zaharescu2012keypoints} point cloud based registration \cite{salti2014shot} and of course recognition and pose estimation \cite{guo2013rotational}.

A very different class of methods rely on 2.5D data, \eg from RGB-D sensors. The best-known method is arguably LINEMOD \cite{hinterstoisser2012gradient}, which allowed for real-time matching of thousands of object templates in RGB-D data. Many competing methods using template-based approaches were introduced afterward, including \cite{brachmann2014learning,rios2013discriminatively,tejani2014latent} and most recently \cite{doumanoglou2016recovering,kehl2016deep,wohlhart2015learning}. The last three achieved very high detection rates by learning an intermediate feature layer with either convolutional or autoencoder neural networks.

Our method bears similarities with the 3D Hough voting of \cite{tombari2010object}, which computed full rotation frames at each feature and used the feature correspondences to cast votes for the 3 DoF translation component of the pose. Although a full rotation can also be computed for each vote, the authors instead performed the mode finding in the reduced 3 DoF Hough space and used RANSAC \cite{fischler1981random} to find the rotation. In contrast, our method casts multiple full 6 DoF votes for each feature correspondence, and we use a branch and bound search strategy on the actual pose samples to find peaks. The method that we use for computing the individual pose votes is in principle similar to that of the point pair features \cite{drost2010model}. In this work randomly sampled oriented point pairs were matched and pose clustering was performed by binning in a low-dimensional pose space. However, while this method used individual point pair correspondences to compute a single pose vote, our method computes a constant number of pose votes for a single point correspondence and delays the inference of the modal pose to the subsequent clustering stage. This difference reduces the complexity of our sampling stage from quadratic to linear.

Finally, some recent methods \cite{aldoma2016global,doumanoglou2016recovering,papazov2010efficient} introduced an additional joint optimization stage. These methods used multiple candidate poses per object and performed a global optimization over the possible combinations of poses to find a configuration of objects that was consistent with the observed scene data. Any pose estimation method, including ours, can in principle be used to provide inputs to these joint optimization frameworks, but the overall performance depends entirely on the ability of the underlying pose estimation algorithms to produce good candidate poses. This ability is the focus of the method presented in this work and the competing methods included in our experiments.

% METHOD FIGURE
\begin{figure*}[ht]
  \centering
    \includegraphics[height=120px]{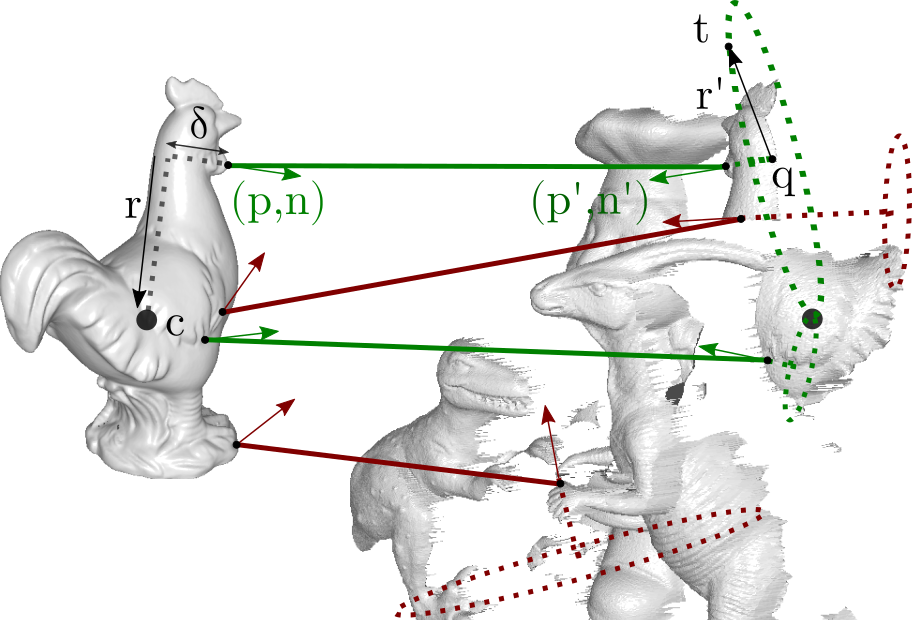}
    \hfill
    \includegraphics[height=120px]{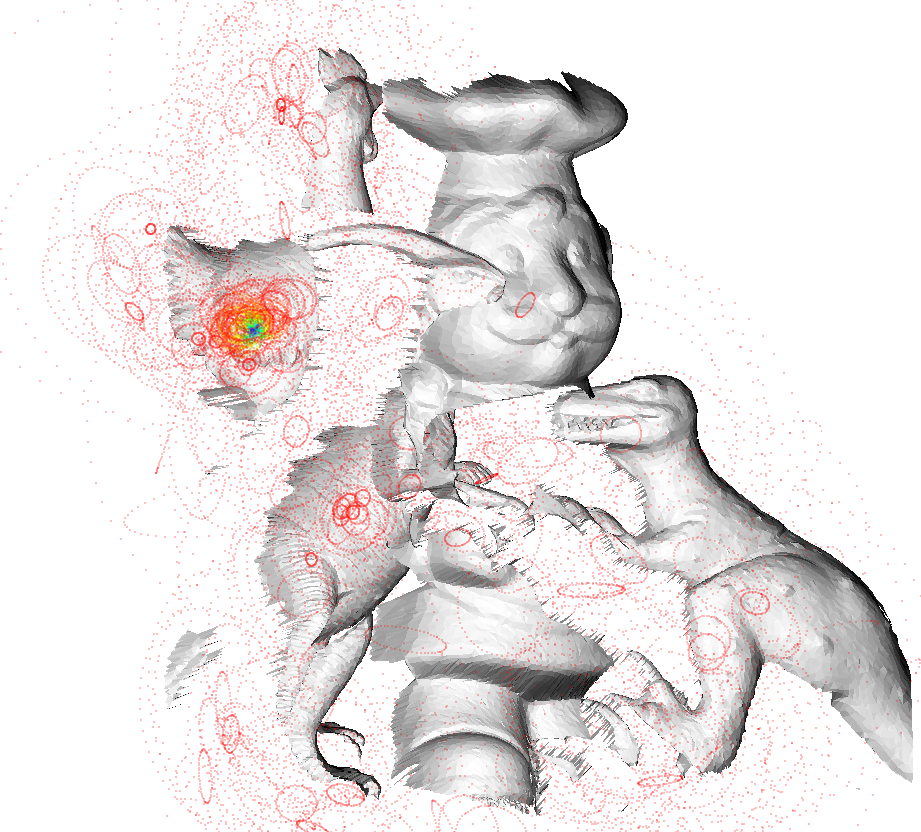}
    \hfill
    \includegraphics[height=120px]{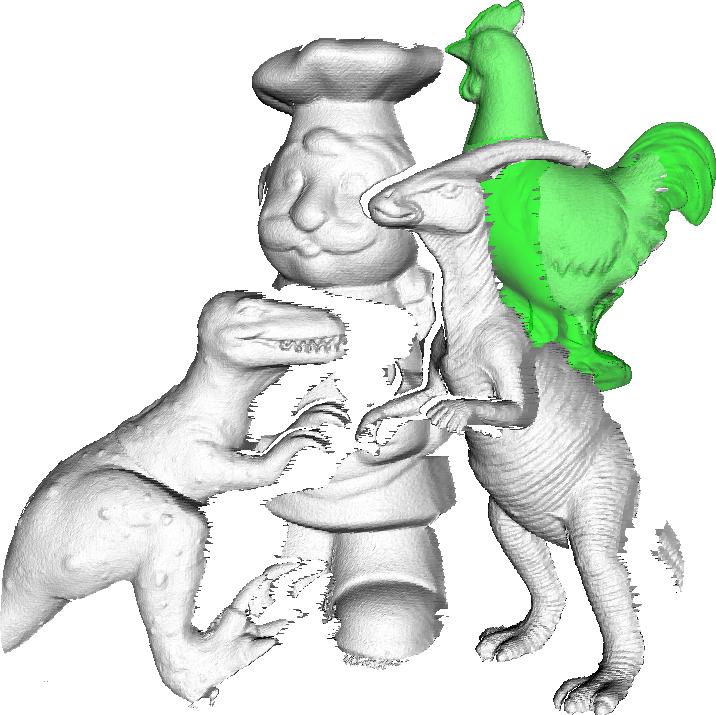}
    \caption{An example scenario visualizing the pose voting process between the Chicken model and the first scene of the UWA data set used for the first test in \refsec{results}. Left: two correct (green) and incorrect (red) correspondences are used to show the geometry of our method. The votes (dashed circles) of the correct correspondences cluster near the center of the object instance in the scene (rightmost black dot). Middle: the same scene---now seen from the back---showing all the votes cast for the model center, with each vote colored proportionally to its $SE(3)$ density estimate (blue is the highest). Right: the final modal pose estimate shown by overlaying the aligned object model in green.}
    \label{fig:voting}
\end{figure*}

%%%%%%%%% METHOD
\section{Method}\label{method}
This section gives the details of our method, which is an algorithm for computing one or more relative 6 DoF poses between one or more 3D object models and a scene. Scene data are usually obtained using a sensor, \eg an RGB-D camera or a laser scanner. In the general case, the challenge is to locate the instance(s) of the object(s) present within the scene, while remaining robust towards inaccuracies (noise), missing data (occlusions) and irrelevant data (cluttering elements). Our algorithm assumes correspondences between surface points on the two models (the object and the scene) are available. We will explain how these correspondences are obtained using local shape features in \refsec{recognition}.

Our algorithm is related to the point pair feature (PPF) method \cite{drost2010model}, as it uses surface normals to define local frames. The PPF method uses a large set of point pair matches between the models to compute full object-relative rotation frames and thereby cast votes for the object pose. Another type of methods \cite{guo2013rotational,tombari2010object} estimates a full local reference frame (LRF) directly at each surface point from the underlying point cloud data. This approach reduces the voting to a linear operation, but results have been suboptimal \cite{tombari2010object}, most likely because the estimation of LRF is unstable in noisy and occluded data. Our algorithm lies in between these two approaches. Similar to the LRF method, we require only a linear number of votes, but we avoid the estimation of a potentially unstable LRF. Instead, for each point match, we compute multiple LRFs using the surface point and the center point of the object. The use of the normal orientation significantly limits the number of possible votes for each point, as the votes are constrained to a 1 DoF manifold. The method is formalized in the following sections.

\subsection{Subgroup voting}
We denote an oriented point on the object model as $(p,n)$, with $p$ being the point coordinates and $n$ being the 3D normal vector pointing away from the surface. A matched feature from the scene provides a correspondence with an oriented point in the scene, which we denote $(p',n')$. In the left part of \reffig{fig:voting} 
\setlength\intextsep{0pt}
\begin{wrapfigure}[4]{l}{0.0875\textwidth}
\includegraphics[width=0.125\textwidth]{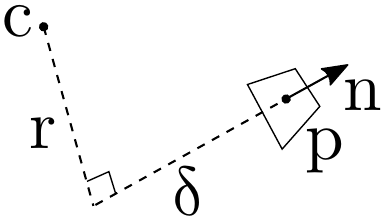}
\end{wrapfigure}
we show some examples of correct matches (green)  and incorrect matches (red). We first compute the scalar projection $\delta$ of the vector going from the object center $c$ to the object point $p$ onto the unit normal $n$:
\begin{equation}
  \delta = (p-c) \cdot n
\end{equation}
We now start from $p$ and follow the negative of $n$ with a distance of $\delta$. Then we compute a \emph{radial vector} $r$ going from this point to the center $c$:
\begin{equation}
  r = c - (p-\delta n)
\end{equation}

\begin{wrapfigure}[4]{l}{0.0875\textwidth}
\includegraphics[width=0.125\textwidth]{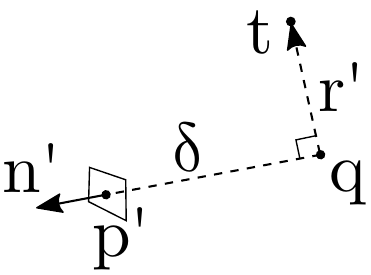}
\end{wrapfigure}
\noindent At this point we have enough information to cast votes for possible translations of the object center into the scene. Using the matched point $(p',n')$ we again follow the negative scene normal using the stored projection for the object center, $\delta$. We call this the \emph{radial point}, $q$:
\begin{equation}
  q = p' - \delta n'
\end{equation}
The voting for the translation of the object center into the scene proceeds as follows. We start by sampling a random vector orthogonal to $n'$ and scale it to a length equal to the radial vector $\|r\|$. Denote this vector $r'$, and note that it is an instantiation of the object radial vector $r$ in the scene.

\begin{wrapfigure}[5]{l}{0.0875\textwidth}
\includegraphics[width=0.125\textwidth]{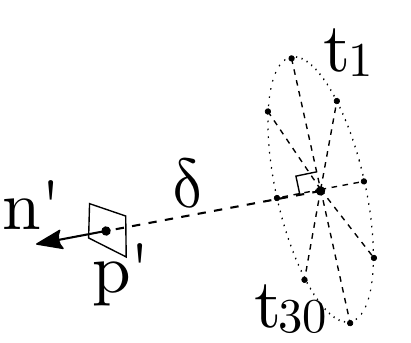}
\end{wrapfigure}
\noindent We now choose a tessellation level $N_r$ and perform incremental rotations of $r'$ around $n'$ with an angle of $\theta=360^\circ/N_r$. Rodrigues' rotation formula can be used to rotate a vector ($r'$) around another ($n'$) by a specified angle ($\theta$) and we use it $N_r$ times to get the \emph{next} instantiation of the radial vector:
\begin{align}
  r' &\leftarrow r'\cos\theta + (n'\times r')\sin\theta + n'(n'\cdot r')(1-\cos\theta) \nonumber\\
     &= r'\cos\theta + (n'\times r')\sin\theta
\end{align}
In our case, $n'$ and $r'$ are always orthogonal, allowing us to eliminate the last term of this equation, as is done in the second line above. For each of the $N_r$ rotated versions of $r'$, we add it to the radial point $q$ and get a candidate translation $t$ of the center of the object into the scene as follows:
\begin{equation}
  t = q + r'
\end{equation}
We refer again to \reffig{fig:voting} for a visualization of the different geometric elements described here.

In the final part of our voting scheme, we show how to recover a full 3 DoF relative rotation for each of the $N_r$ candidate translations $t$. Looking at \reffig{fig:voting}, it can be observed that all the translation candidates lie on a circle. This is a result of the fact that when the corresponding normal vectors $n$ and $n'$---which both have 2 DoFs---are aligned, there is only 1 degree of freedom left to determine. To find this last DoF, we first compute a full 3 DoF rotation frame at the oriented object point $(p,n)$ and do the same for the $N_r$ tessellation points in the scene.
\begin{wrapfigure}[4]{l}{0.0875\textwidth}
\includegraphics[width=0.125\textwidth]{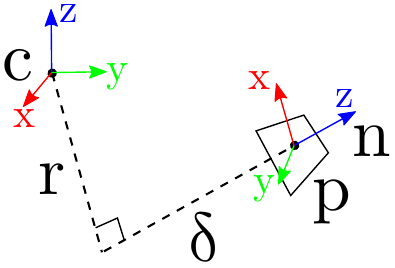}
\end{wrapfigure}
On the object side, this frame is constructed by setting the third column of the rotation frame equal to the normal vector $n$. The radial vector $r$ is always orthogonal to $n$, and we set the first column to this vector normalized. The final vector making up a full rotation frame for the feature point is computed using the cross product. We thus get a 3-by-3 rotation matrix $R_r$ as follows:
\begin{equation}
  R_r = \begin{bmatrix} \frac{r}{\|r\|} & n \times \frac{r}{\|r\|} & n \end{bmatrix}
\end{equation}

\begin{wrapfigure}[5]{l}{0.0875\textwidth}
\includegraphics[width=0.12\textwidth]{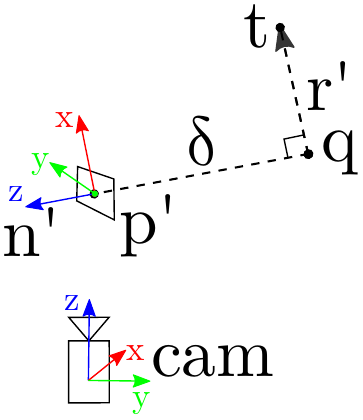}
\end{wrapfigure}
\noindent The same operation is applied to the $N_r$ rotated versions of $r'$ in the scene to get $N_r$ candidate rotation frames $R_{r'}$ as follows:
\begin{equation}
  R_{r'} = \begin{bmatrix} \frac{r'}{\|r'\|} & n' \times \frac{r'}{\|r'\|} & n' \end{bmatrix}
\end{equation}
Finally, the candidate relative rotation $R$ for aligning the object with the scene is given as follows:
\begin{equation}
  R = R_{r'}^\top \cdot R_r
\end{equation}

\begin{wrapfigure}[5]{l}{0.0875\textwidth}
\includegraphics[width=0.125\textwidth]{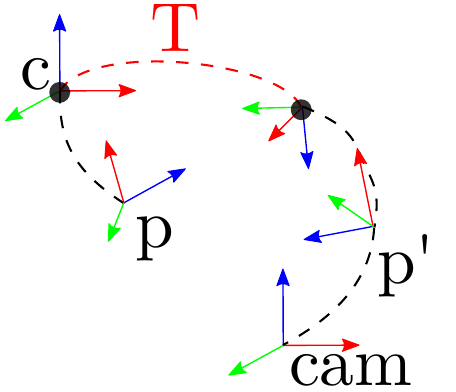}
\end{wrapfigure}
\noindent To summarize, we now have $N_r$ 3 DoF rotations $R$ and 3 DoF translations $t$. Putting each of these together in a 4-by-4 transformation matrix $T$ gives us $N_r$ pose candidates:
\begin{equation}
  T = \begin{bmatrix}   & R &   & t \\
                      0 & 0 & 0 & 1
      \end{bmatrix} \in SE(3)
\end{equation}
For every correct correspondence, there will be one correct and $N_r-1$ incorrect pose votes, all lying on a 1 DoF subgroup of $SO(3)$. All in all, there will a pose vote count equal to $N_r$ times the number of correspondences. We have tried a number of different values for $N_r$ and overall achieved better performance for finer tessellations. We have therefore chosen $N_r=60$ tessellations, giving an angular resolution of $\theta=6^\circ$ of our pose votes. In the following paragraph, we describe how to perform mode finding within these poses to find the correct pose.

\subsection{Density estimation and clustering}
The many pose votes produced by the method described in the previous subsection contain a significant fraction of incorrect candidate poses. However, the six dimensions of the pose group makes it very unlikely that incorrect poses cluster together. This leaves a possibility for the correct poses to cluster together near a detectable mode in $SE(3)$, even though there are only very few of these. This inference process can in principle be performed in different ways, \eg using using $k$-means clustering, mean shift or other mode seeking methods \cite{cheng1995mean}. Unfortunately, the dimensionality of the search space makes many of these approaches intractable due to either excessive memory requirements or high computational complexities.

To overcome this, we use a kernel density estimate on $SE(3)$, computed at each of the pose votes $T$. This requires a measure of distance from each vote to all other votes. A \emph{bandwidth} $\sigma$ is used to preserve locality of the density estimate at each vote. The kernel density estimate $K$ for a pose vote $T$ is computed as follows:
\begin{equation}
  K(T) = \sum_{i=1}^{N_T} f_K(d(T,T_i) / \sigma)
\end{equation}
where $N_T$ is the total number of pose votes, $d$ is some measure of distance between two poses, and $f_K$ is the kernel function. In this work we use the unnormalized Gaussian kernel:
\begin{equation}
  f_K(x) = \exp\left(\frac{-x^2}{2}\right)
\end{equation}
Defining the metric $d$ on $SE(3)$ is non-trivial. Instead, we decompose the density estimate to a product of two 3D Gaussian kernels, one for translations and one for rotations:
\begin{equation}\label{eq:kde}
  K(T) = \sum_{\hat T \in \mathcal{N}(T)} f_K\left(\frac{d_t(\hat t,t)}{\sigma_t}\right) \cdot f_K\left(\frac{d_R(\hat R,R)}{\sigma_R}\right)
\end{equation}
We use the Euclidean distance for the translations:
\begin{equation}\label{eq:euclidean}
  d_t(\hat t,t) = \|\hat t - t\|
\end{equation}
For the rotations we use the minimal geodesic distance along the manifold:
\begin{equation}\label{eq:geodesic}
  d_R(\hat R,R) =  \arccos\left(\frac{\operatorname{trace}(\hat R^\top  R) - 1}{2}\right)
\end{equation}
which lies in the interval $[0,180]^\circ$ and gives the minimal angle needed to rotate $R$ into $\hat R$. In the $SE(3)$ density estimate in \refequ{eq:kde} we changed the summation to occur over a neighborhood around $T$, denoted $\mathcal{N}(T)$. This change is possible because the kernel $f_K$ decays rapidly away from the center. We, therefore, do not need to brute-force loop over all other votes to compute a reliable density estimate for $T$. Instead, we can perform a branch and bound radius search around $T$, with the influence radius set equal to the bandwidth, and find all pose votes within a neighborhood.

To find the neighbor poses $\mathcal{N}(T)$, we first perform a radius search in $\mathbb{R}^3$ using a $k$-d tree to find an initial set of pose neighbors within the translation bandwidth:
\begin{equation}
  \mathcal{N}(t) =  \lbrace \hat t ~ : ~ d_t(\hat t, t) \leq \sigma_t \rbrace
\end{equation}
The full pose neighbors $\mathcal{N}(T)$ are now bounded significantly, since none of these can be outside the set $\mathcal{N}(t)$. We can, therefore, do a linear search \emph{within} $\mathcal{N}(t)$ to find the subset of neighbors within the rotation bandwidth:
\begin{equation}
%  \mathcal{N}(T) = \lbrace (\hat R,\hat t) ~ : ~ d_t(t,\hat t) \leq \sigma_t \wedge d_R(R,\hat R) \leq \sigma_R \rbrace
    \mathcal{N}(T) = \lbrace \hat R \in \mathcal{N}(\hat t) ~ : ~ d_R(\hat R, R) \leq \sigma_R \rbrace
\end{equation}
To summarize, during inference we visit every pose vote, make a search query to find the neighbors, and then visit every neighbor pose and accumulate the density estimate using \refequ{eq:kde}. The poses with high densities represent local modes in $SE(3)$ and provide the output pose estimates $T$ for aligning the object model with one or more instances present in the scene data:
\begin{equation}
  T = \argmax_{\hat T} K(\hat T)
\end{equation}
An alignment using the modal pose is shown in the right part of \reffig{fig:voting}.

\subsection{Computational complexity}\label{complexity}
Like many other competing methods, our algorithm is based on correspondences from local shape features. The computation of features is an $O(N\log N)$ operation in the number of feature points $N$ \cite{muja2014scalable}. The PPF method does not require expensive feature computation, but instead it requires a sampling stage with a complexity of $O(N^2)$. RANSAC does rely on feature correspondences but needs samples with a cardinality of at least three for computing a candidate pose, in which case the complexity rises to $O(N^3)$. The most computationally expensive part of our method is the density estimation stage, where we perform a radius search among all the $SE(3)$ votes. This is an $O(N_r N\log(N_r  N))$ operation, where $N_r$ is the number of rotational tessellations. As mentioned in the previous section, we use a fixed value of $N_r=60$. We thus have a complexity of $O(N\log N)$ with a considerable factor.

% SENSITIVITY RESULTS FIGURE
\begin{figure*}[t]
  \centering
  \hfill{}
  \includegraphics[height=105px]{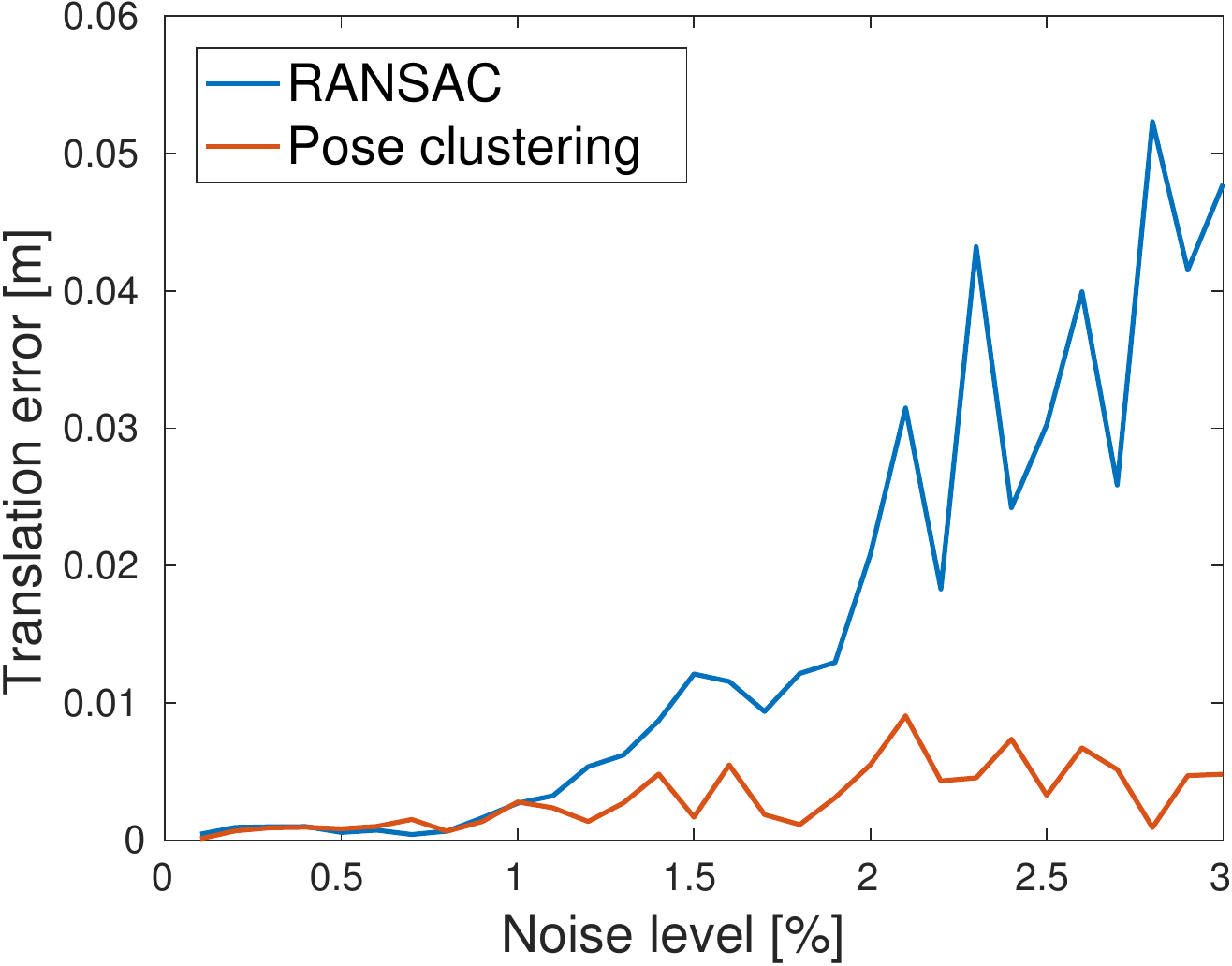}
  \hfill
  \includegraphics[height=105px]{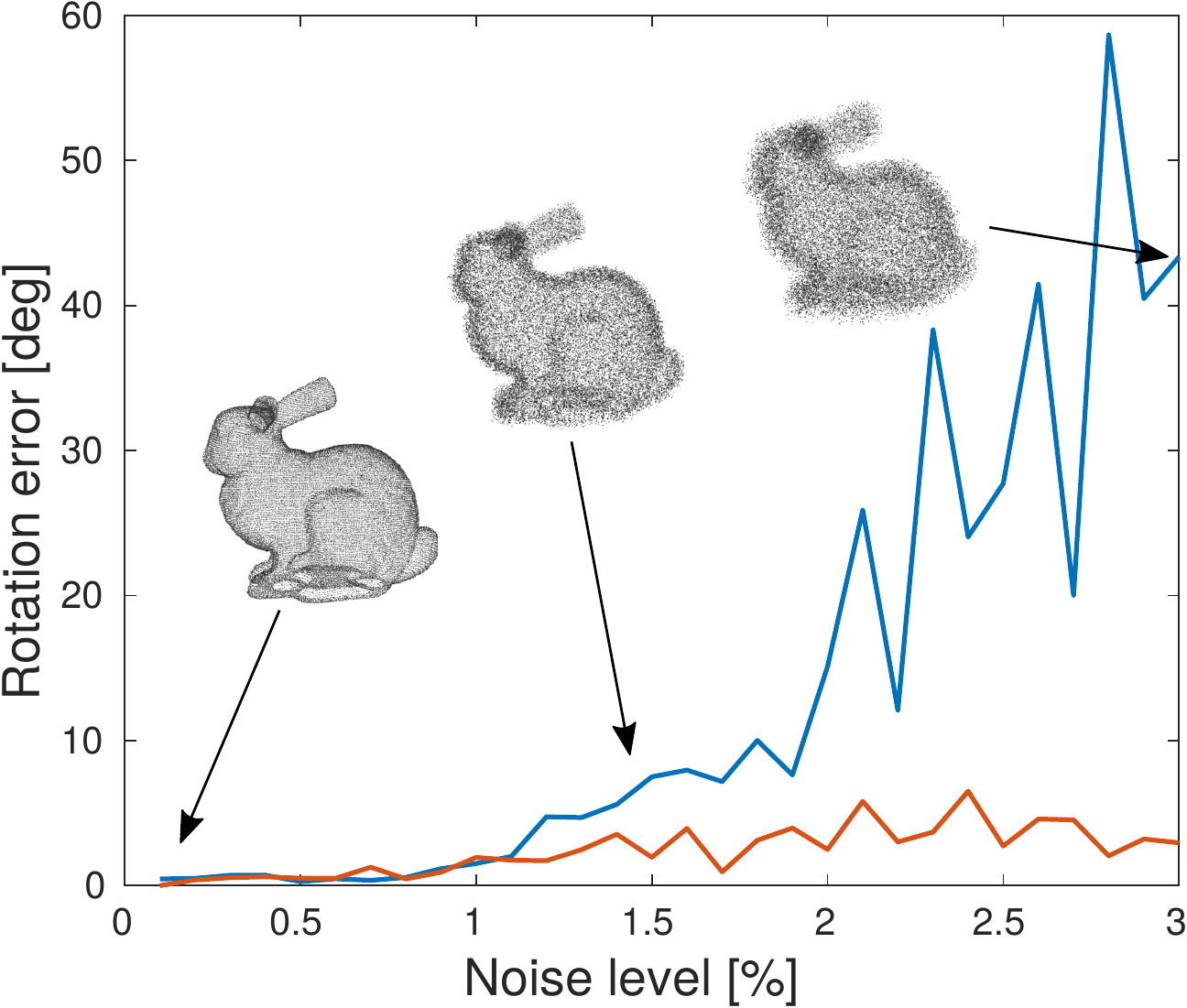}
  \hfill
  \includegraphics[height=105px]{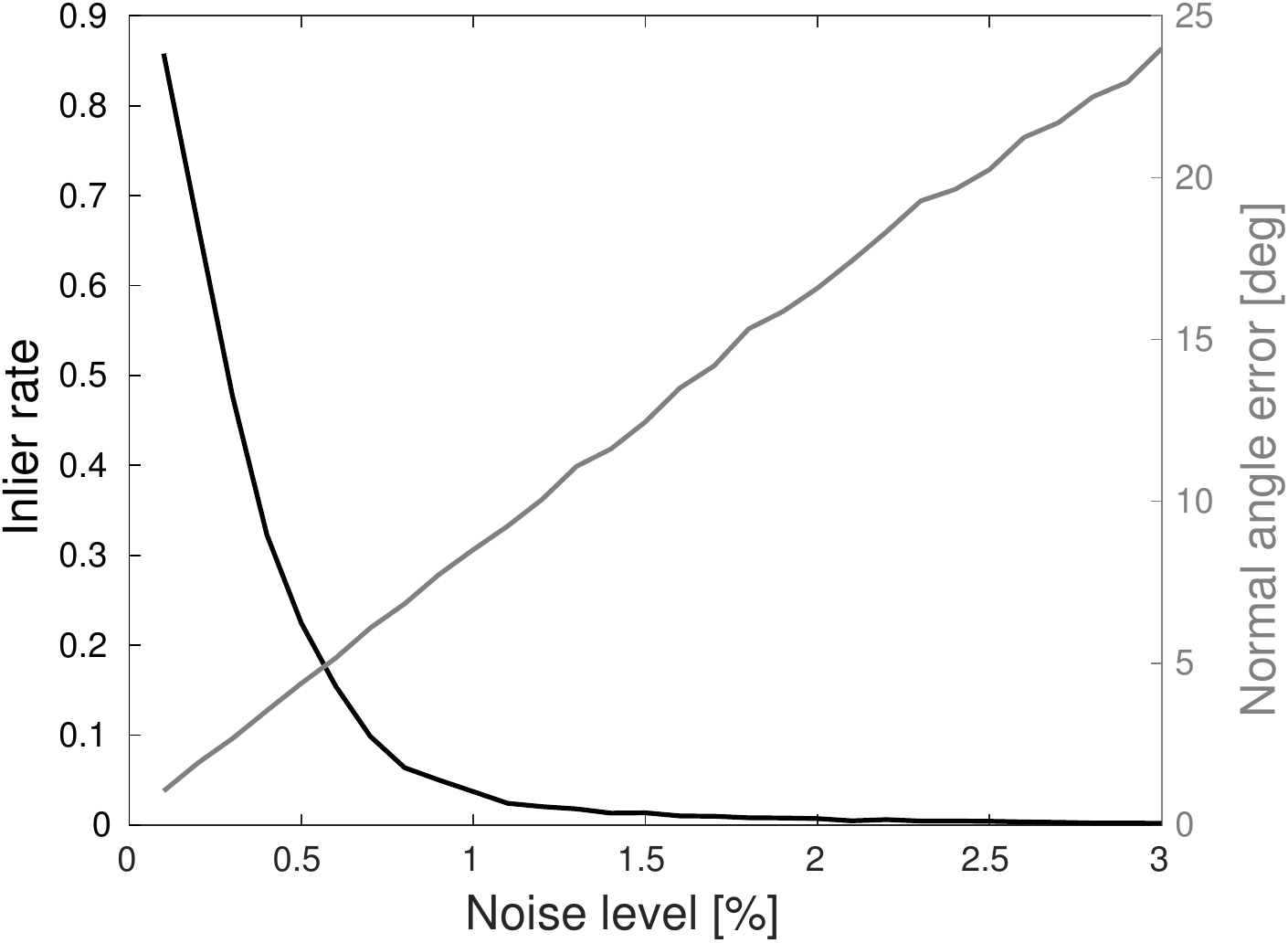}
  \hfill{}
  
  \caption{Sensitivity analysis results obtained by matching the Stanford Bunny to its increasingly noisy counterparts using RANSAC and our method. Left: translation errors for increasing noise levels. Middle: the corresponding rotation errors. Right: ground truth inlier rates for the correspondence sets (black) and average normal angle deviations (gray).}
  \label{fig:sensitivity}
\end{figure*}

%%%%%%%%% PIPELINE
\section{Object recognition and pose estimation pipeline}\label{recognition}
The contributed method detailed in the previous section takes part of a feature-based 3D object recognition pipeline. The input to our pose clustering method is a set of correspondences, which are obtained by matching local features, which again require a good estimate of the local surface normals. When dealing with multiple objects, we process the models sequentially and invoke the pose clustering once for each object. We explain in this section the overall approach taken in our recognition system and end by detailing how we can also use our method for multi-instance recognition of the same object.
%The source code for our method is publicly available in our C++ library\footnote{The link to the library cannot be provided now due to the double blind review process. It will be provided after review.}.
The source code for our method is publicly available in the CoViS C++ library\footnote{\url{https://gitlab.com/caro-sdu/covis}}.

\paragraph{Preprocessing}
Our algorithm can use both triangular meshes or raw point clouds as input models; both types of input data are treated similarly, with the one difference being that for a mesh model we use the faces to ensure globally consistent (i.e. outward-pointing) normals for the object models. For point clouds, we compute surface normals using PCL \cite{rusu20113d} and use a breadth-first search to traverse the object surface and orient the normals consistently. We then downsample both models to a constant resolution using a voxel grid to limit the amount of data for processing. To further reduce the processing time, we avoid computing local features at all surface points but use only a uniform subset of around 10000 feature points per object model.

\paragraph{Feature computation and matching}
To boost the performance of our recognition system, we use our own library for computing discriminative local features. We use randomized $k$-d trees for fast approximate neighbor searches \cite{muja2014scalable} and we always perform search queries with the scene features into an offline generated randomized $k$-d tree index of all the object features.

\paragraph{Multi-object and multi-instance recognition}
The feature matching stage produces a set of correspondences relating the feature points in the scene to points on the objects. In case of multiple objects, we order the objects according to how many correspondences were found for them, starting with the object with most correspondences in the scene. We then run our pose clustering to get the modal pose and use ICP \cite{besl1992method} to refine the estimate. We then adopt a fast, greedy approach, where we segment out the scene data containing the object instance, before proceeding to the next object.

Our pose clustering method also allows for multi-instance recognition, which is tested in the last part of the next section. We first let our pose clustering method return the highest-density pose of each correspondence and then perform non-maximum suppression on this subset of poses using a Euclidean threshold on the translation components equal to 20\,\% of the object model bounding box diagonal.

The next section presents test results for our recognition system in a range of recognition applications.

%%%%%%%%% RESULTS
\section{Results}\label{results}
This section presents five experiments on 3D object recognition and 6 DoF pose estimation. We first present a sensitivity analysis to motivate the use of our method for robust pose estimation under challenging conditions. Then we show results for two well-known data sets, where the objective is to perform recognition and pose estimation for several objects in highly occluded and cluttered real scenes captured by a LIDAR sensor. Finally, we show results for two newer RGB-D data sets, one made for multi-instance detections and one made for detection of domestic objects.

\subsection{Pose clustering as a robust pose estimator}
We first show a sensitivity analysis of our method. For comparison, we included RANSAC \cite{fischler1981random}, which is likely the most robust pose estimator available, cf. its widespread use in 3D recognition, \eg \cite{aldoma2016global,papazov2010efficient,taati2011local}. The task is to align a 3D model to itself under varying noise levels while monitoring the pose errors. We took the classical Stanford Bunny\footnote{\url{http://graphics.stanford.edu/data/3Dscanrep}}, consisting of 35947 points and 69451 triangles, and applied increasing random uniform displacements to the points using MeshLab\footnote{\url{http://meshlab.sourceforge.net}}, with the displacement norm bounded to a specified percentage of the diagonal. For the Bunny, the diagonal is 0.25\,m, and we used noise levels from 0.1\,\% to 3.0\,\% with increments of 0.1\,\%. We computed local features with a spacing of 0.005\,m---giving approximately 3000 feature descriptors---on the original model and each noisy version. We matched the original model to each corrupted model and monitored the ground truth inlier rate by checking how many features on the clean model matched to the same point on the corrupted model with a tolerance of 0.005\,m.

The relative pose to be found here is simply the identity transformation, which allows us to easily measure pose errors. For positions, the error is given by the norm of the estimated translation, and for rotations, we can compute the error as the geodesic distance between the rotation estimate and the identity rotation by \refequ{eq:geodesic}. Contrary to all subsequent experiments, no refinement was used, since the purpose was to investigate the robustness of the two estimators.

The results are shown in \reffig{fig:sensitivity}. For RANSAC, we tuned the number of iterations to 10000 to allow the algorithm to spend approximately the same runtime as ours. For our method, we used a translation bandwidth of 0.01\,m and a rotation bandwidth of 22.5$^\circ$, which are exactly the same parameters that we used in all other experiments. One difference, however, is that since RANSAC is non-deterministic by design, we repeated the the estimation 20 times and took the mean over the 20 runs at each noise level. As shown, our method remains considerably more robust towards noise. We believe the explanation is that our method is better at handling many outliers (wrong correspondences) occurring at high noise levels by virtue of a more discriminating score function. Indeed, our method produces a score proportional to the frequency that a pose occurs in a certain 6 DoF neighborhood of $SE(3)$, which makes spurious local maxima highly accidental. Conversely, RANSAC samples a cubic number of poses and uses a geometric consistency scoring criterion. When the noise increases, there is a much higher risk that many correspondences will support a wrong pose.

The inlier rate, shown rightmost in \reffig{fig:sensitivity}, drops from 86\,\% at the lowest noise level to 0.2\,\% at the highest noise level. Even under such extreme conditions, our method produces the correct pose, whereas RANSAC fails to estimate the relative pose with a translation error of 0.048\,m and a rotation error of 43$^\circ$. Finally, we include a plot of the noise in the surface normals, since our method crucially relies on these for the voting process. We computed the average angular deviation between the normals on each of the corrupted models and those on the original model. As can be seen rightmost in \reffig{fig:sensitivity}, there is a strong linear dependency of these normal deviations on the artificial point noise, leading to the conclusion that our pose estimates are equivalently robust towards noise in the normals. The normals achieve average displacements of almost 25$^\circ$, while our method still produces correct results.

\subsection{Recognition results on the UWA data set}
We first tested our method on the UWA data set \cite{mian2006three}, which is the most well-known and established data set in the literature and has been the subject of several evaluations. The data set contains four complete object models and 50 scenes, all captured with a laser scanner and given as high-resolution triangular meshes. Almost all scenes contain all objects, giving 188 instances to recognize in total. The objects are highly occluded, with a less than 25\,\% average visibility in each scene. We ran the multi-object pipeline outlined in \refsec{recognition} to search for the objects in each scene. We compared against a select number of classical and recent, best-performing methods. For the PPF method, originally proposed in \cite{drost2010model}, we used the latest and optimized implementation of the PPF method, which is now part of the commercial machine vision software Halcon 13.0.0.2\footnote{\url{http://www.mvtec.com/products/halcon}}. We will denote this method PPF* in the folllowing.

Results for the UWA data set are given in the middle column of \reftab{tab:uaw_queens}. For all methods, we give recall rates between 0 and 1, where 1 means 100\,\% recognition rate. Concerning precision, we used a lower threshold on the modal pose density \refequ{eq:kde} to reject false positives and increase precision.

To our knowledge, we are the first to achieve a 100\,\% recognition rate on this data set without the use of joint optimization, as in \eg \cite{aldoma2016global,papazov2010efficient}. Additionally, our method produced few false positives, resulting in a precision of 96.9\,\% and a maximum $F_1$ score of 0.995.

% UWA AND QUEENS RESULTS TABLE
\begin{table}[t]
    \small
    \centering
	  \begin{tabular}[b]{|l|c|c|}
	    \hline
	    Method & UWA & Queen's \\
	    \hline\hline
      Spin Images \cite{johnson1999using} & 0.878 & --- \\

      Tensor matching \cite{mian2006three} & 0.966 & --- \\
            PPF* & 0.936 & 0.992 \\
      %Correspondence voting \cite{buch2014search} & 0.973 & --- \\
      EM \cite{bariya20123d} & 0.975 & 0.824 \\
      %Feature fusion \cite{buch2016local} & 0.984 & 0.913\\
      RoPS \cite{guo2013rotational} & 0.989 & 0.954\\
VD-LSD \cite{taati2011local} & --- & 0.838 \\
	    Pose clustering & \textbf{1.00} & \textbf{1.00} \\
	    \hline
	  \end{tabular}
    \caption{Recall rates for the UWA \cite{mian2006three} and Queen's \cite{taati2011local} data sets. All results except the ones for PPF* and Pose clustering are taken from the literature.}
    \label{tab:uaw_queens}
\end{table}

\subsection{Recognition results on Queen's data set}
The next tested data set was the Queen's data set \cite{taati2011local}, created in a similar manner to the UWA data set. This data set has five objects, 80 scenes and 240 instances. Compared to UWA, it contains a higher variation in the number of objects present in each scene. Each scene also contains spurious data from the ground plane and in general all models are of lower quality and have non-uniform resolution.

We report comparative results for the Queen's data set in the right column of \reftab{tab:uaw_queens}, which reveals reduced recognition rates for many of the competing methods relative to UWA. Remarkably, both our method and PPF* perform better than UWA on this data set. PPF* achieves a recall of 99.2\,\% and the same precision. On this data set, our method achieves 100\,\% recall at a precision of 100\,\%.

\subsection{Recognition results on the bin picking data set}
\begin{figure}[t]
    \centering
    \includegraphics[height=150px]{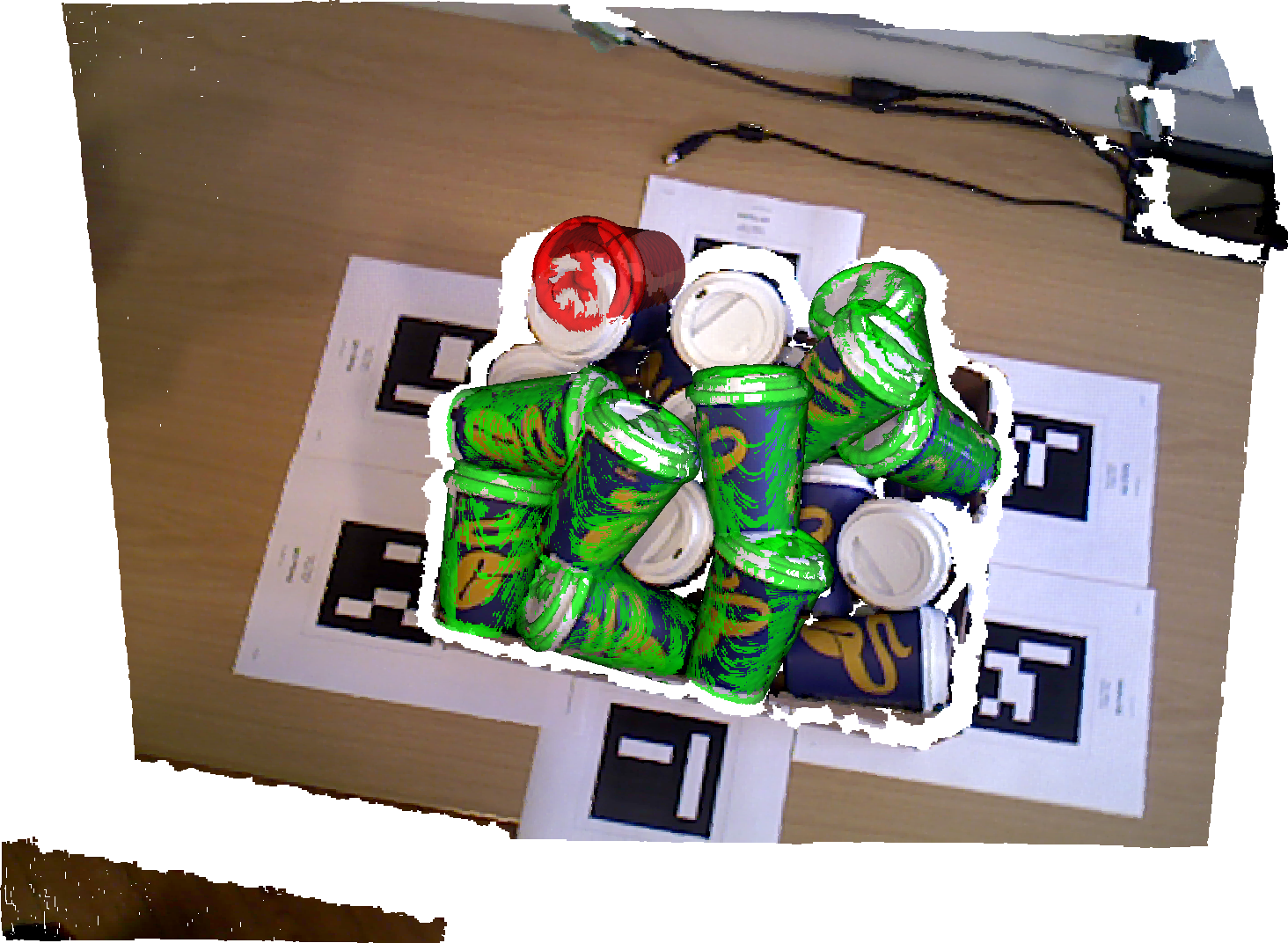}
    \caption{Multi-instance recognition output (the top ten detections) for the first bin picking scene with nine true positives (green) and one false positive in the back (red).}
    \label{fig:icl}
\end{figure}

Another experiment was done on a very recent data set \cite{doumanoglou2016recovering}, where the authors introduced a bin picking data set consisting of 183 RGB-D images showing multiple instances of two test objects in a small bin. The scenes are split up in three sequences: one where the bin contains 15 instances of the Coffee cup object, one where the bin contains five instances of the Juice box object and finally a mixed sequence where each image shows the bin containing nine and four instances of the Coffee cup and the Juice box, respectively. The protocol for this data set is to match the two objects to their dedicated sequences and to the mixed sequence. Although this data set is primarily targeted at another class of detection methods---namely RGB-D based systems that use both color and depth information---we wanted to test the performance of our method, even though our method relies purely on geometric cues. On the other hand, the PPF* method and ours do not require expensive training but derive the features for matching directly from the oriented surfaces. Contrary to the two previous experiments, the objective is now multi-instance detection, so we extracted the ten top ranked modes after non-maximum suppression using our method. The same applies to the PPF* method where we set it to return the ten top scoring poses. Other than that, all experimental parameters for both methods were the same as previously. The results, including the baseline results from \cite{doumanoglou2016recovering,tejani2014latent}, are given in \reftab{tab:icl} and a multi-instance recognition example is shown in \reffig{fig:icl}.

% BIN PICKING RESULTS TABLE
\begin{table}[t]
    \small
    \centering
    \begin{tabular}[b]{|l|c|c|}
      \hline
      Method & Object & Recall \\
      \hline\hline
      \multirow{2}{*}{Tejani \etal \cite{tejani2014latent}}%
      & Coffee cup & 0.314 \\
      & Juice & 0.248 \\
      \hline
      \multirow{2}{*}{Doumanoglou \etal \cite{doumanoglou2016recovering}}%
      & Coffee cup & 0.335 \\
      & Juice & 0.251 \\
      \hline
      \multirow{2}{*}{PPF*}%
      & Coffee cup & 0.474 \\
      & Juice & 0.279 \\
      \hline
      \multirow{2}{*}{Pose clustering}%
      & Coffee cup & \textbf{0.638} \\ % March'17
      & Juice & \textbf{0.449} \\ % March'17
      \hline
    \end{tabular}
    \caption{Recall rates for the bin picking data set \cite{doumanoglou2016recovering}.}
    \label{tab:icl}
\end{table}

% BIN PICKING EXAMPLE FIGURE

The results show that the 3D methods (PPF* and ours) compete well with RGB-D based methods. This is achieved in approximately the same runtime for all methods, which is in the order of seconds per recognized instance. These results support the use of our method for multi-instance recognition problems. We believe the main reason why our method is able to outperform PPF* is that we use smooth density estimates on $SE(3)$, whereas PPF* uses approximate clustering. Our method achieves a substantial improvement over other methods, in particular the two RGB-D based methods, which were designed for this kind of data.

\subsection{Recognition results on domestic data set}
The final experiment was performed on the data set of~\cite{tejani2014latent}, which is a challenging RGB-D based recognition data set for domestic environments, containing thousands of test scenes. Results have been reported for LINEMOD~\cite{hinterstoisser2012gradient}, PPF~\cite{drost2010model} and two new RGB-D based methods~\cite{doumanoglou2016recovering,tejani2014latent}. As recommended in~\cite{tejani2014latent}, we extract the top five modes in each scene to build the precision-recall curves. For all objects except the Shampoo, our method produces a higher $F_1$ score than the other 3D method, PPF. For the Camera object, the most recent RGB-D method \cite{doumanoglou2016recovering} outperforms ours. On average, our method outperforms existing methods, producing the highest average $F_1$ score. The results are listed in~\reftab{tab:tejani}.

We stress that the methods \cite{doumanoglou2016recovering,hinterstoisser2012gradient,tejani2014latent} use both geometric and appearance cues from RGB-D templates, whereas PPF and our method use only the geometry to match a full 3D model to a scene view. We would like to try using color-based local 3D features with our approach, as this should allow for further improvements for RGB-D data sets.

% TEJANI RESULTS TABLE
\begin{table}[t]
    \small
    \centering
    \begin{tabular}[b]{|l|c|c|c|c|c|}
      \hline
      %Object & LINEMOD [21] & PPF [16] & Tejani \etal [39] & \thead{Doumanoglou \\ \etal [14]} & \thead{Pose\\clustering} \\
      Object & \cite{hinterstoisser2012gradient} & \cite{drost2010model} & \cite{tejani2014latent} & \cite{doumanoglou2016recovering} & Ours \\
      \hline\hline
      Coffee cup & 0.819 & 0.867 & 0.877 & 0.932 & \textbf{0.993} \\
      Shampoo    & 0.625 & 0.651 & \textbf{0.759} & 0.735 & 0.709 \\ % March'17
      Joystick   & 0.454 & 0.277 & 0.534 & 0.924 & \textbf{0.973} \\
      Camera     & 0.422 & 0.407 & 0.372 & \textbf{0.903} & 0.711 \\ % March'17
      Juice      & 0.494 & 0.604 & 0.870 & 0.819 & \textbf{0.975} \\
      Milk       & 0.176 & 0.259 & 0.385 & 0.510 & \textbf{0.776} \\
      \hline
      Average    & 0.498 & 0.511 & 0.633 & 0.803 & \textbf{0.856} \\
      \hline
    \end{tabular}
    \caption{$F_1$ scores for the data set \cite{tejani2014latent} and the following methods: LINEMOD~\cite{hinterstoisser2012gradient}, PPF~\cite{drost2010model}, Tejani \etal~\cite{tejani2014latent}, Doumanoglou \etal~\cite{doumanoglou2016recovering} and our pose clustering method.}
    \label{tab:tejani}
\end{table}

\subsection{Runtimes}
In practice, our system has a per-object runtime---including preprocessing, feature computation and matching, which are all amortized over all objects---of 2--4\,s (3.5\,s for UWA, 4\,s for Queen's, 2\,s for the bin picking data set and 3\,s for the domestic data set). These numbers are obtained by execution on a consumer laptop with a 2.60\,GHz Intel i7-5600U CPU with four cores, leaving a potential for speedup on other architectures and with further optimizations. More than 90\,\% of this time is spent on 3D features and matching to obtain correspondences. Thus, the clustering and mode finding is not the bottleneck in our system.

% AVG. PER-INSTANCE RECOGNITION TIMES
%   UWA:3.5 s
%   Queen's: 4s
%   Bin picking: 2 s
%   Tejani: 3 s

For PPF* we used a regularly updated implementation from the Halcon software (we used v.\ 13.0.0.2), which also runs in a few seconds per instance. In \cite{doumanoglou2016recovering}, the total processing time is unspecified, although it is stated that the main bottleneck of the system takes 4--7\,s. In \cite{tejani2014latent} the runtime is unspecified. Other systems relying on local features report runtimes such as a few seconds \cite{aldoma2016global} and minutes \cite{guo2013rotational,mian2006three}.

%%%%%%%%% CONCLUSION
\section{Conclusions and future work}\label{conclusions}
This work contributed a method for 3D object recognition using a new pose voting and clustering method for obtaining robust pose estimates in cluttered scenes. The pose voting exploited the fact that corresponding oriented points between two models can be used to cast a constrained number of votes for the correct pose aligning the two models. For the final inference step, a branch and bound search was performed to compute density estimates for each pose. An initial sensitivity analysis showed increased robustness to outlier correspondences compared to RANSAC. When integrated into a local feature-based recognition pipeline, our method achieved perfect recall for two well-known recognition data sets and it has outperformed recent methods on two RGB-D recognition data sets.

Our method is slightly sensitive towards planar or repetitive structures, since the local feature correspondences scatter randomly in their presence. We are currently investigating whether other local features, \eg edge based, can be used to obtain better correspondences under such conditions. We are also working on incorporating appearance information into our method using color-based local features, which should allow for increased accuracy on RGB-D data.

%%%%%%%%% ACKNOWLEDGEMENTS
\section*{Acknowledgements}
The research leading to these results has been funded in part by Innovation Fund Denmark as a part of the project \enquote{MADE --- Platform for Future Production} and by the EU FoF Project ReconCell (project number 680431).
%\clearpage

{\small
\bibliographystyle{ieee}
\bibliography{references}

\begin{thebibliography}{10}\itemsep=-1pt

\bibitem{aldoma2016global}
A.~Aldoma, F.~Tombari, L.~Di~Stefano, and M.~Vincze.
\newblock A global hypothesis verification framework for 3d object recognition
  in clutter.
\newblock {\em IEEE Transactions on Pattern Analysis and Machine Intelligence},
  38(7):1383--1396, 2016.

\bibitem{bariya20123d}
P.~Bariya, J.~Novatnack, G.~Schwartz, and K.~Nishino.
\newblock 3d geometric scale variability in range images: Features and
  descriptors.
\newblock {\em International Journal of Computer Vision}, 99(2):232--255, 2012.

\bibitem{besl1992method}
P.~Besl and N.~D. McKay.
\newblock A method for registration of 3-d shapes.
\newblock {\em IEEE Transactions on Pattern Analysis and Machine Intelligence},
  14(2):239--256, 1992.

\bibitem{birdal2015point}
T.~Birdal and S.~Ilic.
\newblock Point pair features based object detection and pose estimation
  revisited.
\newblock In {\em IEEE International Conference on 3D Vision}, pages 527--535,
  2015.

\bibitem{brachmann2014learning}
E.~Brachmann, A.~Krull, F.~Michel, S.~Gumhold, J.~Shotton, and C.~Rother.
\newblock Learning 6d object pose estimation using 3d object coordinates.
\newblock In {\em European Conference on Computer Vision}, pages 536--551,
  2014.

\bibitem{chen20073d}
H.~Chen and B.~Bhanu.
\newblock 3d free-form object recognition in range images using local surface
  patches.
\newblock {\em Pattern Recognition Letters}, 28(10):1252--1262, 2007.

\bibitem{cheng1995mean}
Y.~Cheng.
\newblock Mean shift, mode seeking, and clustering.
\newblock {\em IEEE Transactions on Pattern Analysis and Machine Intelligence},
  17(8):790--799, 1995.

\bibitem{chua1997point}
C.~S. Chua and R.~Jarvis.
\newblock Point signatures: A new representation for 3d object recognition.
\newblock {\em International Journal of Computer Vision}, 25(1):63--85, 1997.

\bibitem{doumanoglou2016recovering}
A.~Doumanoglou, R.~Kouskouridas, S.~Malassiotis, and T.-K. Kim.
\newblock Recovering 6d object pose and predicting next-best-view in the crowd.
\newblock In {\em IEEE Conference on Computer Vision and Pattern Recognition},
  June 2016.

\bibitem{drost20123d}
B.~Drost and S.~Ilic.
\newblock 3d object detection and localization using multimodal point pair
  features.
\newblock In {\em Second International Conference on 3D Imaging, Modeling,
  Processing, Visualization \& Transmission}, pages 9--16, 2012.

\bibitem{drost2010model}
B.~Drost, M.~Ulrich, N.~Navab, and S.~Ilic.
\newblock Model globally, match locally: Efficient and robust 3d object
  recognition.
\newblock In {\em IEEE Conference on Computer Vision and Pattern Recognition},
  volume~1, page~5, 2010.

\bibitem{fischler1981random}
M.~A. Fischler and R.~C. Bolles.
\newblock Random sample consensus: a paradigm for model fitting with
  applications to image analysis and automated cartography.
\newblock {\em Communications of the ACM}, 24(6):381--395, 1981.

\bibitem{frome2004recognizing}
A.~Frome, D.~Huber, R.~Kolluri, T.~B{\"u}low, and J.~Malik.
\newblock Recognizing objects in range data using regional point descriptors.
\newblock In {\em European Conference on Computer Vision}, pages 224--237,
  2004.

\bibitem{guo20143d}
Y.~Guo, M.~Bennamoun, F.~Sohel, M.~Lu, and J.~Wan.
\newblock 3d object recognition in cluttered scenes with local surface
  features: a survey.
\newblock {\em IEEE Transactions on Pattern Analysis and Machine Intelligence},
  36(11):2270--2287, 2014.

\bibitem{guo2013rotational}
Y.~Guo, F.~Sohel, M.~Bennamoun, M.~Lu, and J.~Wan.
\newblock Rotational projection statistics for 3d local surface description and
  object recognition.
\newblock {\em International Journal of Computer Vision}, 105(1):63--86, 2013.

\bibitem{hinterstoisser2012gradient}
S.~Hinterstoisser, C.~Cagniart, S.~Ilic, P.~Sturm, N.~Navab, P.~Fua, and
  V.~Lepetit.
\newblock Gradient response maps for real-time detection of textureless
  objects.
\newblock {\em IEEE Transactions on Pattern Analysis and Machine Intelligence},
  34(5):876--888, 2012.

\bibitem{hinterstoisser2016going}
S.~Hinterstoisser, V.~Lepetit, N.~Rajkumar, and K.~Konolige.
\newblock Going further with point pair features.
\newblock In {\em European Conference on Computer Vision}, pages 834--848,
  2016.

\bibitem{johnson1999using}
A.~E. Johnson and M.~Hebert.
\newblock Using spin images for efficient object recognition in cluttered 3d
  scenes.
\newblock {\em IEEE Transactions on Pattern Analysis and Machine Intelligence},
  21(5):433--449, 1999.

\bibitem{kehl2016deep}
W.~Kehl, F.~Milletari, F.~Tombari, S.~Ilic, and N.~Navab.
\newblock Deep learning of local rgb-d patches for 3d object detection and 6d
  pose estimation.
\newblock In {\em European Conference on Computer Vision}, pages 205--220,
  2016.

\bibitem{mian2006three}
A.~S. Mian, M.~Bennamoun, and R.~Owens.
\newblock Three-dimensional model-based object recognition and segmentation in
  cluttered scenes.
\newblock {\em IEEE Transactions on Pattern Analysis and Machine Intelligence},
  28(10):1584--1601, 2006.

\bibitem{muja2014scalable}
M.~Muja and D.~G. Lowe.
\newblock Scalable nearest neighbor algorithms for high dimensional data.
\newblock {\em IEEE Transactions on Pattern Analysis and Machine Intelligence},
  36(11):2227--2240, 2014.

\bibitem{papazov2010efficient}
C.~Papazov and D.~Burschka.
\newblock An efficient ransac for 3d object recognition in noisy and occluded
  scenes.
\newblock In {\em Asian Conference on Computer Vision}, pages 135--148, 2010.

\bibitem{rios2013discriminatively}
R.~Rios-Cabrera and T.~Tuytelaars.
\newblock Discriminatively trained templates for 3d object detection: A real
  time scalable approach.
\newblock In {\em IEEE International Conference on Computer Vision}, pages
  2048--2055, 2013.

\bibitem{rusu20113d}
R.~B. Rusu and S.~Cousins.
\newblock 3d is here: Point cloud library (pcl).
\newblock In {\em IEEE International Conference on Robotics and Automation},
  pages 1--4, 2011.

\bibitem{salti2014shot}
S.~Salti, F.~Tombari, and L.~Di~Stefano.
\newblock Shot: unique signatures of histograms for surface and texture
  description.
\newblock {\em Computer Vision and Image Understanding}, 125:251--264, 2014.

\bibitem{taati2011local}
B.~Taati and M.~Greenspan.
\newblock Local shape descriptor selection for object recognition in range
  data.
\newblock {\em Computer Vision and Image Understanding}, 115(5):681--694, 2011.

\bibitem{tejani2014latent}
A.~Tejani, D.~Tang, R.~Kouskouridas, and T.-K. Kim.
\newblock Latent-class hough forests for 3d object detection and pose
  estimation.
\newblock In {\em European Conference on Computer Vision}, pages 462--477,
  2014.

\bibitem{tombari2010object}
F.~Tombari and L.~Di~Stefano.
\newblock Object recognition in 3d scenes with occlusions and clutter by hough
  voting.
\newblock In {\em Fourth Pacific-Rim Symposium on Image and Video Technology},
  pages 349--355, 2010.

\bibitem{wohlhart2015learning}
P.~Wohlhart and V.~Lepetit.
\newblock Learning descriptors for object recognition and 3d pose estimation.
\newblock In {\em IEEE Conference on Computer Vision and Pattern Recognition},
  pages 3109--3118, 2015.

\bibitem{zaharescu2012keypoints}
A.~Zaharescu, E.~Boyer, and R.~Horaud.
\newblock Keypoints and local descriptors of scalar functions on 2d manifolds.
\newblock {\em International Journal of Computer Vision}, 100(1):78--98, 2012.

\end{thebibliography}
}

\end{document}